\documentclass{article}

\PassOptionsToPackage{round}{natbib}

\usepackage[final]{nips_dlps_2017}

\usepackage[utf8]{inputenc} % allow utf-8 input
\usepackage[T1]{fontenc}    % use 8-bit T1 fonts
\usepackage{hyperref}       % hyperlinks
\usepackage{url}            % simple URL typesetting
\usepackage{booktabs}       % professional-quality tables
\usepackage{amsfonts}       % blackboard math symbols
\usepackage{nicefrac}       % compact symbols for 1/2, etc.
\usepackage{microtype}      % microtypography
\usepackage{graphicx}
\usepackage{color}
\usepackage{float}
\usepackage{subfigure}
\usepackage{authblk}
\usepackage{tikz}
\usepackage{amsmath}       % blackboard math symbols
\usetikzlibrary{shapes.geometric}
\usetikzlibrary{positioning}

\title{Variational Inference over Non-differentiable Cardiac Simulators using Bayesian Optimization}

\author[1]{Adam McCarthy}
\author[1]{Blanca Rodriguez}
\author[1]{Ana Minchol\'e}

\affil[1]{
  Department of Computer Science\\
  University of Oxford\thanks{\texttt{firstname.lastname@cs.ox.ac.uk}}}

\begin{document}
% \nipsfinalcopy is no longer used
\vspace{-1.5em}
\maketitle
\vspace{-1.5em}
\begin{abstract}
	Performing inference over simulators is generally intractable as their runtime means we cannot compute a marginal likelihood. We develop a likelihood-free inference method to infer parameters for a cardiac simulator, which replicates electrical flow through the heart to the body surface. We improve the fit of a state-of-the-art simulator to an electrocardiogram \emph{(ECG)} recorded from a real patient.
\end{abstract}
\section{Introduction}

The heart contracts due to the propagation of electricity through the heart wall. This electricity then propagates through the torso and we can record an electrocardiogram \emph{(ECG)}. Cardiac simulators replicate this propagation, outputting an ECG mirroring that recorded from a real patient \citep[see][for a review]{Cardone-Noott2016}. They are slow, require significant compute power - generally supercomputers, and are non-differentiable (we cannot compute its jacobian analytically). We must fit the parameters of these simulators efficiently so that they replicate the recording taken from real patients and provide clinicians with deep, interpretable understandings of the sources of risk. These insights also aid basic research and the development of new treatments.

This combination of slow run time and lack of differentiability make parameter inference and uncertainty propagation significant challenges and active areas of research. We cannot compute a likelihood due to the simulator run time and these problems frequently have many minima.

An approach to parameter inference is variational inference, which provides a parametric approximation of the posterior of a probabilistic model. Early work on variational inference was limited to forms in which gradient descent steps could be computed analytically \citep[e.g.][]{Jordan1999}. Nowadays it is possible to take steps where the closed form is not known, for example using the \emph{reparameterisation trick} \citep{Kingma2014a}. This uses a Monte Carlo estimate of the likelihood to approximate the gradient of the optimisation function, but requires that the simulator is differentiable. \cite{Ranganath2014} drop this requirement with the \emph{score function estimate} but  gradient estimates are noisy. We focus on the optimiser instead of the gradients by using Bayesian optimisation, which being a global optimisation method makes converging to local minima less likely.

This problem of parameter inference over non-differentiable simulators applies to many fields in which simulators are used. We need to be able to fit weather simulators to predict hurricanes \citep{Rappaport2009}, simulators of the formation of the universe \citep{Boylan-Kolchin2014}, and simulators to predict the spread of wildland fires \citep{Sullivan2009}, to give a few examples. We also need to understand how changes in these parameters affect the output, and how certain the simulators are that events will happen.

\paragraph{Contribution} We develop a method to infer parameters for a non-differentiable simulator using Variational Inference combined with Bayesian Optimisation, allowing us to fit a posterior distribution with a minimal number of trials and avoid local minima where possible. We show that we can use the inferred parameters to improve the fit of a state-of-the-art cardiac simulator.
\vspace{-1em}
\begin{figure}[H]
    \centering
    \begin{tikzpicture}[square/.style={regular polygon,regular polygon sides=4}]
            \node  [circle, draw] (theta) {$\theta$};
            \node [square, draw, right = 1.5cm of theta] (heart) {$h$};
            \node  [circle, draw, right = 1.5cm of heart] (ecg) {$ECG$};

            \draw [->] (theta) edge (heart) (heart) edge (ecg);
            \draw [->] (ecg) to [out=150,in=30] (theta);
    \end{tikzpicture}
    \caption[The inverse problem via inference]{The forward problem of the ECG involves taking the anatomy as a prior and defining parameters representing biological features, then propagating electricity through the heart, diffusing the charge through the torso and producing a simulated ECG. The inverse problem involves performing inference over the forward problem. $\theta$ are the model parameters, $h$ is a heart simulator, and $ECG$ is the output ECG trace.}
    \label{fig:flowchart}
\end{figure}
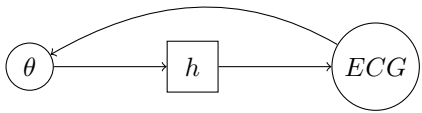

\section{Methodology}
\label{sec:methodology}
\paragraph{Simulator} The state-of-the-art cardiac simulator implements the bidomain equations \citep[see][for a review]{Trayanova2011}. The bidomain-based simulator, whilst physiologically very accurate, takes several hours to run on a supercomputer. We developed a fast cardiac simulator which includes an anatomical heart model derived from MRI as a prior, simulates electrical propagation through the heart to the body surface, and outputs a 12-lead ECG (see Figure\ref{fig:flowchart} for a visualisation). Whilst this simulator is much faster, it is still too slow to compute a likelihood. It has a 17 dimensional parameter space consisting of locations at which electricity enters the heart wall and propagation velocities through the inner and outer heart wall, which have previously been measured in-vivo. For a biophysical meaning of these parameters, we refer the reader to \cite{Cardone-Noott2016}.

The simulator we use will be presented in future work, so for the purposes of this paper can be a function $f(\theta)$ which has parameters $\theta \in \mathbb{R}^p$ and returns output $X \in \mathbb{R}^d$. It is a deterministic, black box function, with no simple closed form, and that it can be evaluated at any $\theta$ within the domain of interest.  We have observations $ECG \in \mathbb{R}^d$ recorded from the patient and the aim is to compute a posterior $p(\theta|ECG)$. We also test the inferred parameters on the bidomain-based simulator, which can be defined similarly, to demonstrate that we have improved the fit of the state-of-the-art.

\paragraph{Variational inference} Given that our simulator is non-differentiable we cannot use methods which use second-order information \citep[e.g.][]{Regier2017} or reduce the variance of the gradients \citep[e.g.][]{Miller2017,Roeder2017}. We construct our variational objective following the \emph{variational autoencoder} \citep{Kingma2014a} and optimise over the marginals of our Gaussian approximating distribution: $\mu$ and $\sigma$. Variational autoencoders define a parametric generative model with likelihood $p(X|\theta)$ for each dimension, prior $p(\theta)$ over the parameters and a model $q(\theta|X)$ which approximates the posterior. It can be shown that
\begin{align}\label{eq:elbo}
\log p(x) \ge -KL\left(q(\theta|X\right), p(\theta)) + \mathbb{E}_{q(\theta|X)}\left[\log p(X|\theta)\right]
\end{align}

where the right hand term in Equation~\ref{eq:elbo} is termed the \emph{variational lower bound} or \emph{evidence based lower bound objective (ELBO)}. In variational inference we maximise the ELBO by optimising over the marginal parameters of $Q$.

Our approach shares many similarities with the variational autoencoder, but the neural network is replaced by a cardiac simulator. We avoid more recent work on improving the posterior distributions \cite[e.g.][]{Ranganath2016,Liu2016b,Huszar2017} and we will explain the reasons for this in Section~\ref{sec:results}.

\paragraph{Optimisation of the variational objective}

Our simulator is non-differentiable, so we cannot perform gradient-based optimisation without estimating gradients in some way. We could approximate the jacobian using finite differences, but this requires $n$ evaluations of the model, where $n$ is the number of dimensions over which we are optimising. There is also work on approximating gradients in variational inference, but they tend to be noisy. Instead, we perform global optimisation using Bayesian Optimisation to make our method applicable to all non-differentiable simulators.

It is common with local optimisation to grow the KL-divergence over time such that the log likelihood, which fits the mean of $Q$, dominates initially. Global optimisation routines have memory, however, so we fit the log likelihood first and then the variance via the KL-divergence as independent optimisations to prevent the KL-divergence dominating initially.

When performing Bayesian optimisation, we found a Square Exponential kernel was optimal, likely because the optimisation surface should be relatively smooth. We initially used Expected Improvement as our acquisition function, but found that the noisy samples meant Augmented EI \citep{Huang2006} had superior performance. We sample our initial design matrix using Latin Hypercube Sampling and define our parameter space $\theta$ as the endocardial conductivities representing the Purkinje system, three conductivities for anisotropic conductivity in the myo- and epi-cardium, and the position of the stimulus locations, which includes all of the nodes in the inner heart wall \emph{(the endocardium)}. We allow the conductivities to vary $\pm$50\%. For a biophysical meaning of these parameters we refer the reader to \cite{Trayanova2011}. We set bounds on the parameters to represent the underlying physiological variability \citep{Britton2013}, and define inequality constraints for:

\begin{enumerate}
\item The order of three of the conductivities representing the direction of muscle fibres. We know a priori that they are ordered.
\item The stimulus positions are position invariant, i.e. $a,b,c \equiv c,b,a$, so we specify that the first dimension of the latent space representing the positions on a manifold through the heart (which we will discuss in the following paragraph) is ordered.
\end{enumerate}

\paragraph{Dimensionality reduction}

So far we have assumed that our simulator is a black box. In this section, we move to a white box setting and use our knowledge of its internal dynamics to reduce dimensionality. We take magnetic resonance imaging (MRI) scans of the patient's heart and use this to derive an anatomical mesh. When we use this mesh within the simulator by converting it to a graph representation of electrical propagation, we are enforcing a strong prior on its dynamics. The heart itself lies on a lower dimensional manifold within Cartesian space due to the hollow ventricles, which makes optimisation over Cartesian coordinates discontinuous. We reduce the dimensionality of the graph vertices using an \emph{isomap} \citep{Tenenbaum2000}. This works by computing geodesic distances between the vertices, and applying PCA to produce an embedding. We then use this embedding as a 2D manifold over which to optimise and will show in Section~\ref{sec:results} that this significantly improves convergence time. Bayesian optimisation can struggle in large dimensions because the acquisition function, Expected Improvement, becomes flat near minima. We found that reducing the dimensionality improved the accuracy of the inferred parameters.

Isomaps are injective, meaning there is no inverse mapping from the 2D latent space back into Cartesian space. There are, however, a finite number of nodes in the mesh. We compute a lookup table between the nodes in Cartesian space and those in the latent space. We use a kd-tree to take a node in latent space and find its nearest neighbour in Cartesian space using the lookup table. This is necessary to transfer parameters inferred to the bidomain-based simulator, which uses a finer mesh.

\section{Results and discussion}\label{sec:results}

In this section we will report our results and then discuss how they improve fitting these simulators.

\paragraph{Dimensionality reduction} Figure~2a shows the heart mesh reduced to 2D using an isomap over 16 neighbours for each point, which produces a mean reconstruction error of only 0.0016mm when projecting back into Cartesian space. Reducing dimensionality from three to two will have a minimal effect on the speed of fitting of a Gaussian Process, so the improvement in convergence speed and performance (see Figure~2b) is likely related to the increased smoothness of the optimisation surface created by following the manifold.

\paragraph{Parameter fitting} Figure~3 shows an ECG recorded from a real patient and the associated bidomain simulation - the current state-of-the-art. Whilst the bidomain equations are physiologically accurate, their long runtime makes parameter fitting difficult. We take the parameters inferred on the simpler simulator and show that they improve the fit of the bidomain-based simulator which is considered the gold standard in cardiac modelling.

Structural modelling of the anatomy enforces a strong prior on the model which is essential in a small data regime such as medicine. Rather than treating the inverse problem as one of regression \citep[e.g.][]{Intini2005}, we perform inference over a simulator representing the forward problem which provides this prior.

In-vivo measurements vary widely in the literature, due to noisy biological readings. Whilst maintaining uncertainty is important, it should be regarded with some scepticism due to the inaccuracy of the underlying readings on which these models are based. Choosing simple approximating distributions is important for two reasons: firstly they simplify the problem, allowing us to fit based on limited data and secondly they provide simpler, actionable decision support to clinicians. Projecting a Gaussian onto the heart surface is far more interpretable for a surgeon, for example, than a more complex distribution which might span multiple parts of the heart. Enforcing simplicity is an important consideration in the medical domain.

Finally, simulators of this form often present many local minima, so whereas variational inference usually relies on forms of stochastic gradient descent, we have used global optimisation which is more effective for providing accurate effective decision support.

\paragraph{Related work} Approximate Bayesian Computation (ABC) is a form of likelihood-free inference commonly used for these forms of problem. \cite{Tavare1997} were the first to apply ABC, but it was named later by \cite{Beaumont2002}. The simplest form of ABC, \emph{rejection ABC}, is inefficient because $\theta$ is arbitrarily sampled from its prior distribution. leading to numerate rejections. \emph{MCMC ABC} builds a markov chain through the prior, creating a proposal distribution from which to obtain new query points \cite[see][for a review]{Marjoram2003}. Whilst this is more efficient than rejection ABC, it would still be intractable in our case due to the number of samples required.

\cite{Papamakarios2016} avoid rejecting samples by directly approximating the posterior as a mixture of Gaussians: a proposal prior $q(\theta)$. They draw samples from this prior and run the corresponding simulations, iteratively updating the proposal prior using a \emph{mixture density network} (MDN) \citep{Bishop1994}, until it matches the target posterior.

Another approach is to approximate the likelihood function using a parametric model rather than avoiding its computation entirely, a \emph{synthetic likelihood}. \cite{Wood2010} used a single Gaussian distribution and \cite{Fan2012} used a mixture of Gaussians by learning multiple models based on repeated evaluations for fixed values of $\theta$. \cite{Meeds2014} used a Gaussian process to combine the likelihood approximations for each $\theta$. Finally, we note a relationship with \cite{Louppe2017}, which has very similar motivations but relies on local optimisation using Variational Inference combined with a GAN.

In the modelling literature, \cite{VanDam2016} provide a method for non-invasive mapping of the heart from the 12-lead ECG. They enforce a strong prior by using a cardiac simulator as we do, but use local optimisation which may fall into local minima, whereas we use global optimisation. They find a point estimate of the PVC origin, whereas we provide a distribution over possible PVC origins. This is useful clinically, because it gives the clinician a measure of how accurate the model believes the localisation to be.

Lastly, \cite{Giffard-Roisin2017} approach the problem by learning a regression. They also provide uncertainty estimates and perform dimensionality reduction. In contrast to our method, they use 205 sensors across the torso \citep[ECGI body surface potential maps][]{Ramanathan2004}, whereas we take standard 12-lead ECG as input. We can use less input data because the simulator enforces a strong prior. Their optimisation is local whereas our is global and they use Relevance Vector Machines, which are patented by Microsoft, so this may be a consideration when applying their work.

\section{Conclusion}

We have demonstrated an efficient, generic method for parameter inference over cardiac simulators by using Variational Inference and Bayesian Optimization. We have used a cardiac simulator to infer parameters to reproduce the ECG of a patient and used these parameters to improve the fit of a simulator based on the gold standard for cardiac modelling: the bidomain equations.

The ability to do efficient parameter inference makes it possible to personalise simulators and capture much more broadly the underlying electrophysiological variability. We plan to use this work to improve our cardiac simulators by capturing new forms of variability, which will represent a significant improvement in the accuracy of these simulators. We hope researchers from other fields will find similar benefits.

\newpage
\paragraph{Acknowledgements}
Adam McCarthy is supported by EPSRC studentship OUCL/2016/AM. Blanca Rodriguez (BR) and Ana Minchol\'e are supported by BR’s Wellcome Trust Senior
Research Fellowship in Basic Biomedical Sciences, the CompBiomed project (grant agreement
No 675451) and the NC3R Infrastructure for Impact award (NC/P001076/1). Compute infrastructure was provided by the UK National
Supercomputing Service (ARCHER Leadership Award e462) and a Microsoft Azure Research Award. We wish to thank Michael Osborne, Nando de Freitas, and Matthew Graham for helpful discussions, and Andrew Trask for a thorough proofread of our paper.

\begin{figure}[H]
    \centering
    \includegraphics[width=0.41\textwidth]{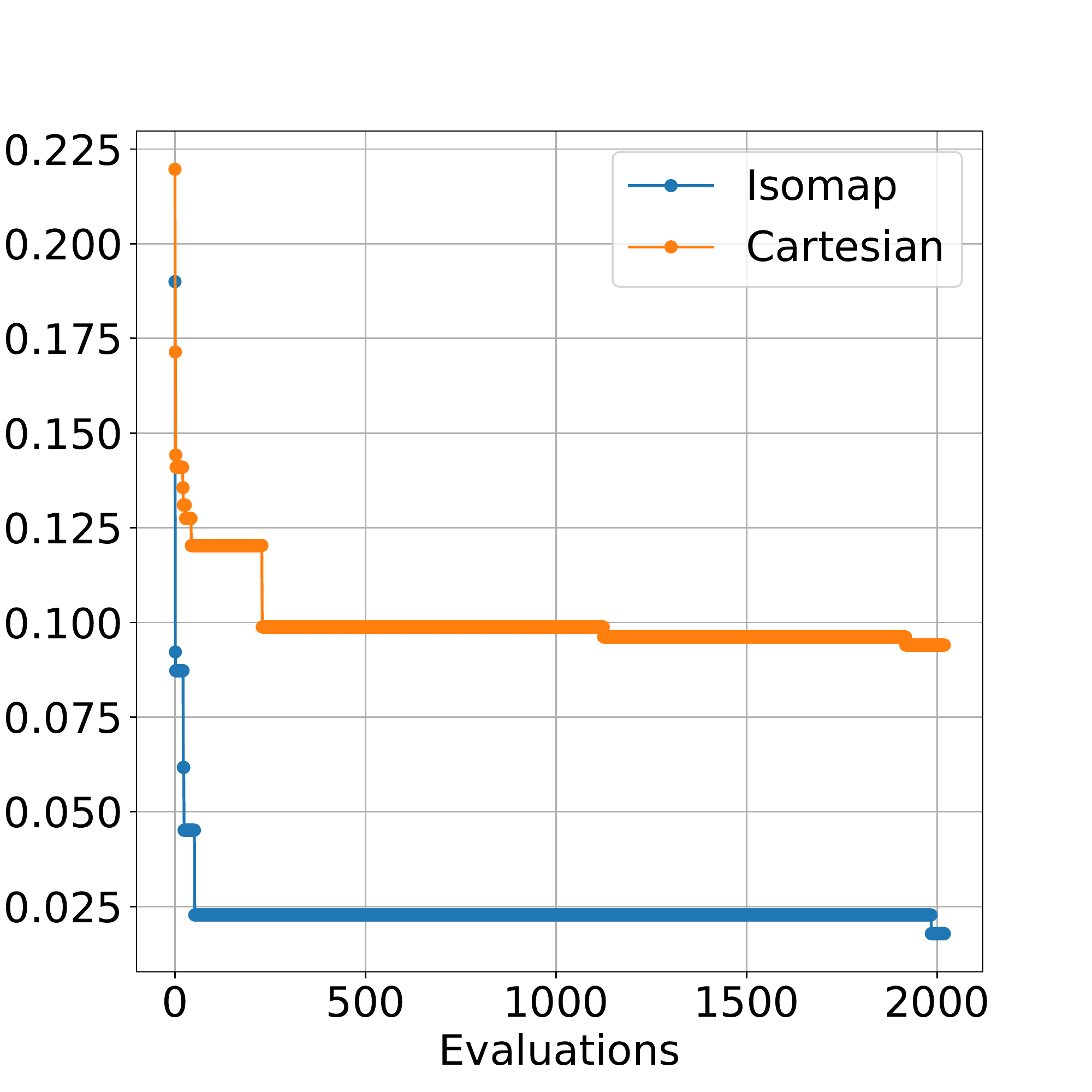}
    \hfill
    \includegraphics[width=0.41\textwidth]{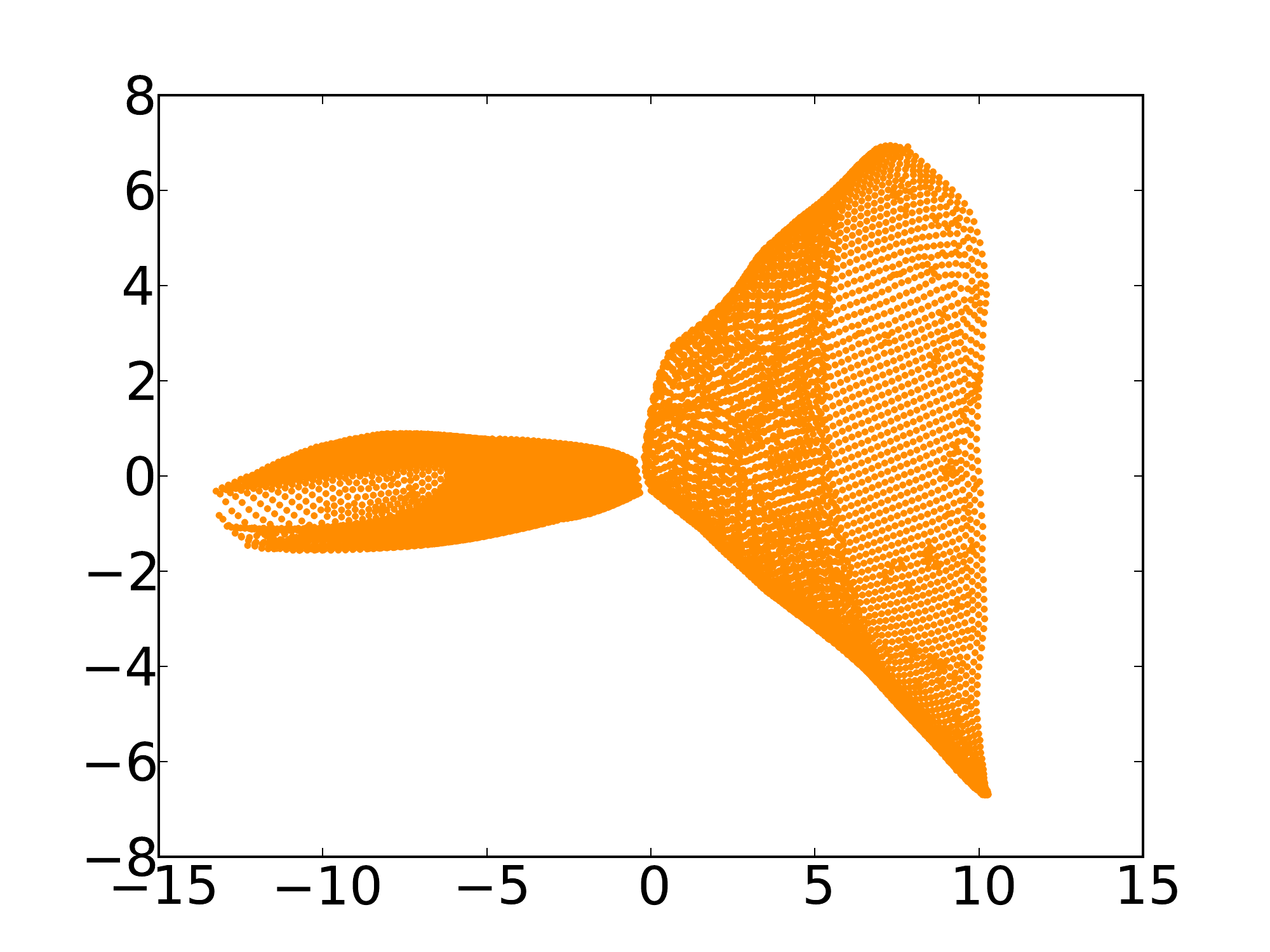}

    \begin{minipage}{0.49\textwidth}\label{fig:cost}
        \centering
        a) MSE cost curves for optimising over cartesian space and a manifold derived from an embedding of the anatomical mesh using an isomap
    \end{minipage}
    \hfill
    \begin{minipage}{0.49\textwidth}\label{fig:embed}
        \centering
        b) A 2D embedding of the inner heart wall, \emph{the endocardium}, using an isomap over 16 neighbours
    \end{minipage}
    \caption{Plots to demonstrate dimensionality reduction}
\end{figure}

\begin{figure}[H]\label{fig:comparison}
    \centering
    \includegraphics[width=0.23\textwidth]{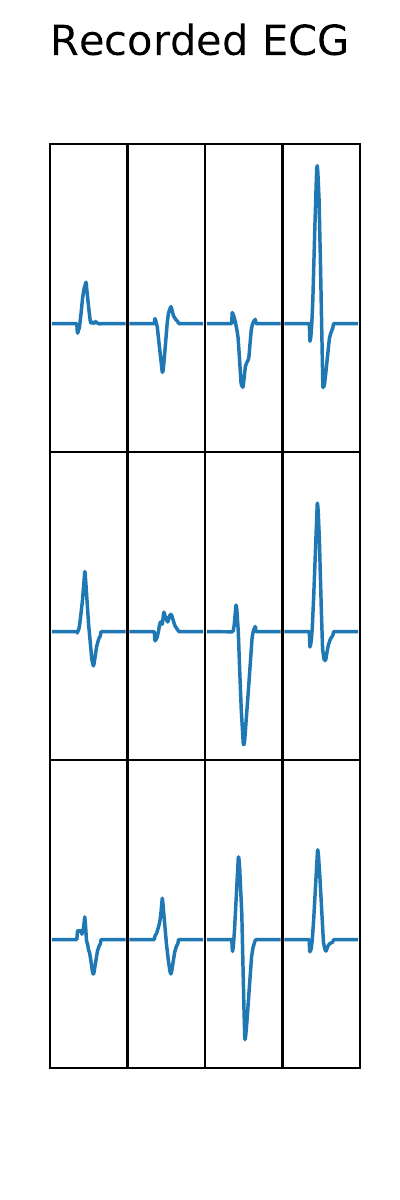}
    \includegraphics[width=0.23\textwidth]{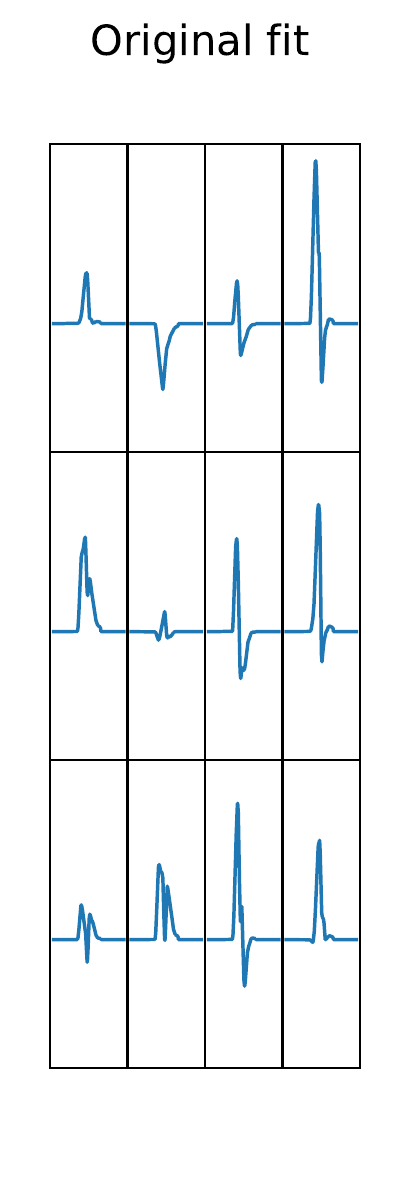}
    \includegraphics[width=0.23\textwidth]{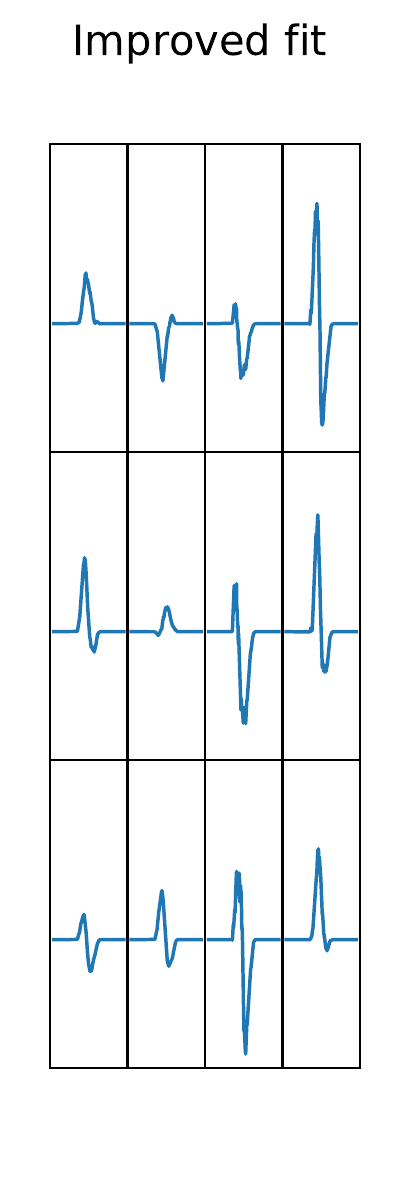}
    \caption{Comparison between the recording from the real patient, the baseline simulation - the current state-of-the-art, and our improved fit. Our inference method allowed us to find parameters which better fit the dynamics of the real patient's heart. The recorded ECG is subject to electrical noise and noise from lead placement.}
\end{figure}

\newpage
\bibliographystyle{abbrvnat}
{\small \bibliography{main}}

\end{document}